\documentclass[%
 reprint,
 amsmath,amssymb,
 aps,
]{revtex4-1}

\usepackage{graphicx}% Include figure files

\begin{document}

\preprint{APS/123-QED}

\title{Convolutional neural networks for classification and regression analysis of one-dimensional spectral data}% Force line breaks with \\

\author{Ine L. Jernelv}
 \email{ine.jernelv@ntnu.no}
\author{Dag Roar Hjelme}%
\author{Astrid Aksnes}
\affiliation{%
 Department of Electronics, Norwegian University of Science and Technology (NTNU), O.S. Bragstads plass 2A, 7491 Trondheim, Norway
}%


\author{Yuji Matsuura}
\affiliation{
Graduate School of Biomedical Engineering, Tohoku University, Sendai, 980-8579, Japan
}%
%\affiliation{
%\affiliation{%


\date{\today}% It is always \today, today,

\begin{abstract}
Convolutional neural networks (CNNs) are widely used for image recognition and text analysis, and have been suggested for application on one-dimensional data as a way to reduce the need for pre-processing steps. Pre-processing is an integral part of multivariate analysis, but determination of the optimal pre-processing methods can be time-consuming due to the large number of available methods. In this work, the performance of a CNN was investigated for classification and regression analysis of spectral data. The CNN was compared with various other chemometric methods, including support vector machines (SVMs) for classification and partial least squares regression (PLSR) for regression analysis. The comparisons were made both on raw data, and on data that had gone through pre-processing and/or feature selection methods. The models were used on spectral data acquired with methods based on near-infrared, mid-infrared, and Raman spectroscopy. For the classification datasets the models were evaluated based on the percentage of correctly classified observations, while for regression analysis the models were assessed based on the coefficient of determination (R$^2$). Our results show that CNNs can outperform standard chemometric methods, especially for classification tasks where no pre-processing is used. However, both CNN and the standard chemometric methods see improved performance when proper pre-processing and feature selection methods are used. These results demonstrate some of the capabilities and limitations of CNNs used on one-dimensional data. 
%\begin{description}
%\end{description}
\end{abstract}

\maketitle

\section{Introduction}
Chemometric methods are an essential part of understanding and interpreting vibrational spectral data. Vibrational spectroscopy uses variants of near-infrared (NIR), mid-infrared (MIR), or Raman spectroscopy techniques, and produce one-dimensional spectral data. Models for spectroscopy data usually try to either map the spectra to distinct classes (classification), or try to extract quantitative information (regression). Designing optimal models is challenging for both classification and regression analysis. Most samples have several constituents with overlapping bands, which makes simple considerations based on e.g. peak heights insufficient for any data interpretation or accurate analysis. Additionally, spectroscopy is used for applications in many different fields, including arts, forensics, food science, environment, and medicine \cite{thygesen2003, Kendall2009, Das2011, Muro2015, Pozzi2016}.

For classification, common chemometric methods include linear discriminant classifiers, support vector machines, $k$-nearest neighbours, and logistic regression methods. For quantitative analysis, regression methods such as principal component regression, partial least-squares regression, and random forest regression are commonly used. Prior to the actual analysis, pre-processing methods and feature selection are often applied to the datasets.

Pre-processing methods are commonly applied to spectral data for several purposes, including noise removal and removal of unwanted physical phenomena that affect the data. The end goal of the applied pre-processing steps is to improve the interpretability of the data, and thereby achieve higher accuracy in the following classification or quantitative multivariate analysis. There are many methods for pre-processing, and the optimal choice of pre-processing methods usually depends on the measurement method used (NIR, MIR, Raman, etc.) and the aim of the analysis \cite{Rinnan2009, Engel2013}. In addition, factors such as sample matrix, equipment settings, and environmental influences can affect the data, so that previously applied pre-processing methods may not work well on a new dataset. Practically, the selection of methods is frequently based on previous experiences, or alternatively an exhaustive search of all possible methods that the researcher has available. Exhaustive searches should give the best results, but are computationally expensive and time-consuming. It has been shown that optimal - or at least close to optimal - pre-processing methods can be selected based on design of experiments \cite{Gerretzen2015a}, which could make the selection much faster.

Spectral data typically has relatively few samples (10s to a few 100s), each with many features (up to 1000s). This can be detrimental to multivariate analysis, as random correlations and noisy features may lead to a deterioration of the model, and a tendency to overfit can give low predictive ability. Feature selection is performed in order to remove irrelevant or redundant features, preferentially leaving only features that are relevant to the analysis \cite{Mehmood2012}. The objective of feature selection is the same as for pre-processing, namely increased interpretability of the data and better outcomes in multivariate analysis. Feature selection can be very computationally expensive and a plethora of feature selection methods exist, while some learning algorithms have built-in feature selection.

One alternative to optimisation of pre-processing and feature selection is to use learning models that are capable of extracting fundamental information from data, for example convolutional neural netoworks (CNNs). Artificial neural networks (ANNs) are computational models inspired by the biological neural connections in brains. One-layer ANNs have been used for classification and regression of spectral data for almost three decades, for example in combination with PLS \cite{Bhandare1993}. ANNs are particularly useful for non-linear problems and can be used on almost any type of data. However, models such as PLS-ANN have not superseded standard chemometric methods, largely due to issues with overfitting and low interpretability \cite{Marini2008}. CNNs can improve upon this, as the convolutional layers of the network are not fully connected, and are therefore in theory less prone to overfitting. Additionally, the convolution leads to fewer free parameters, and each layer becomes related to specific parts of the input data, which increases ease of operation and interpretability.

CNNs are widely used for data mining of two-dimensional data, including areas such as image recognition and text analysis. Neural networks have several interesting characteristics for data modelling, such as the capability to model nonlinear responses and to accentuate relevant features. CNNs can therefore be expected to perform well on learning tasks also for one-dimensional data, even without pre-processing or feature extraction methods. Thus far, only a few studies have used CNNs for classification or quantitative analysis on vibrational spectroscopic data. Acquarelli et al. \cite{Acquarelli2017} demonstrated that CNNs could outperform several standard chemometric methods for classification, both using raw and pre-processed spectral data. CNNs have also been shown to efficiently classify Raman spectra of mineral data without spectral preprocessing \cite{Liu2017}. For quantitative analysis CNNs have been used both for feature selection for regression methods \cite{Malek2018}, and directly for regression analysis \cite{Cui2018}. Outside of standard vibrational spectroscopy, some recent studies have also used CNNs for analysis of other spectral data, such as classification of hyperspectral data \cite{Chen2016}, functional near-infrared spectroscopy signals \cite{Rosas-Romero2019}, and electrocardiogram (ECG) signals \cite{Kiranyaz2016}.

This study will investigate the performance of a CNN as compared to standard chemometric methods for classification and quantitative analysis on several datasets acquired with different spectroscopic methods. For classification the CNN will be compared to PLS-DA, $k$NN, SVM, and logistic regression, while for quantitative analysis the CNN will be compared to PCR, PLSR, random forest regression, and elastic net. The comparison will be made on models both with and without the use of pre-processing methods, as well as with and without feature selection methods. All methods used in this study were assembled in a software package, named SpecAnalysis, which has been made available online. In our view, there are two main original contributions in this work: Use of CNNs for both classification and regression analysis, with a comparison to several standard chemometric methods, and comparison of models with additional use of pre-processing and/or feature selection, to further assess the performance of CNNs.

\section{Materials and methods}

\subsection{Data analysis}
The analysis was done with a software package made in Python, called SpecAnalysis, which can be found on GitHub (\url{https://github.com/jernelv/SpecAnalysis}). Python is an open-source programming language, and can be used on all common operating systems. SpecAnalysis has a graphic user interface, and is therefore user-friendly even for spectroscopists who are not experienced programmers. 

SpecAnalysis has functionality for spectral pre-processing, feature selection, various chemometric and machine learning methods, and can be employed for both regression analysis and classification tasks. Some methods in SpecAnalysis were based on tools from scikit-learn \cite{Pedregosa2011}, while others were made in-house. 

The methods used in this study for classification, regression analysis, and feature selection are briefly described below, and an overview of the pre-processing methods is given. More information on the methods can be found in the references or in the SpecAnalysis documentation. 

\subsubsection{Convolutional neural network}
The basic structure of artificial neural networks (ANNs) consists of connected artificial neurons. Each neuron is characterised by an activation function, which is a function that acts on the input. Neural networks have three types of layers: input, hidden, and output layers. For the input and output layers, which are the first and last layers respectively, the activation functions are generally linear.  Neural networks have one or more hidden layers, where the activation functions are normally non-linear. The output at each layer is used as the input of the next layer. In a fully-connected feed-forward ANN, all neurons in each layer are connected to all neurons in the next layer, with no connections going backward to previous layers. The connections between the layers are weighted, and these weights are learned in the training phase after a random initialisation.
\begin{figure*}
 \centering
 \includegraphics[height=9cm]{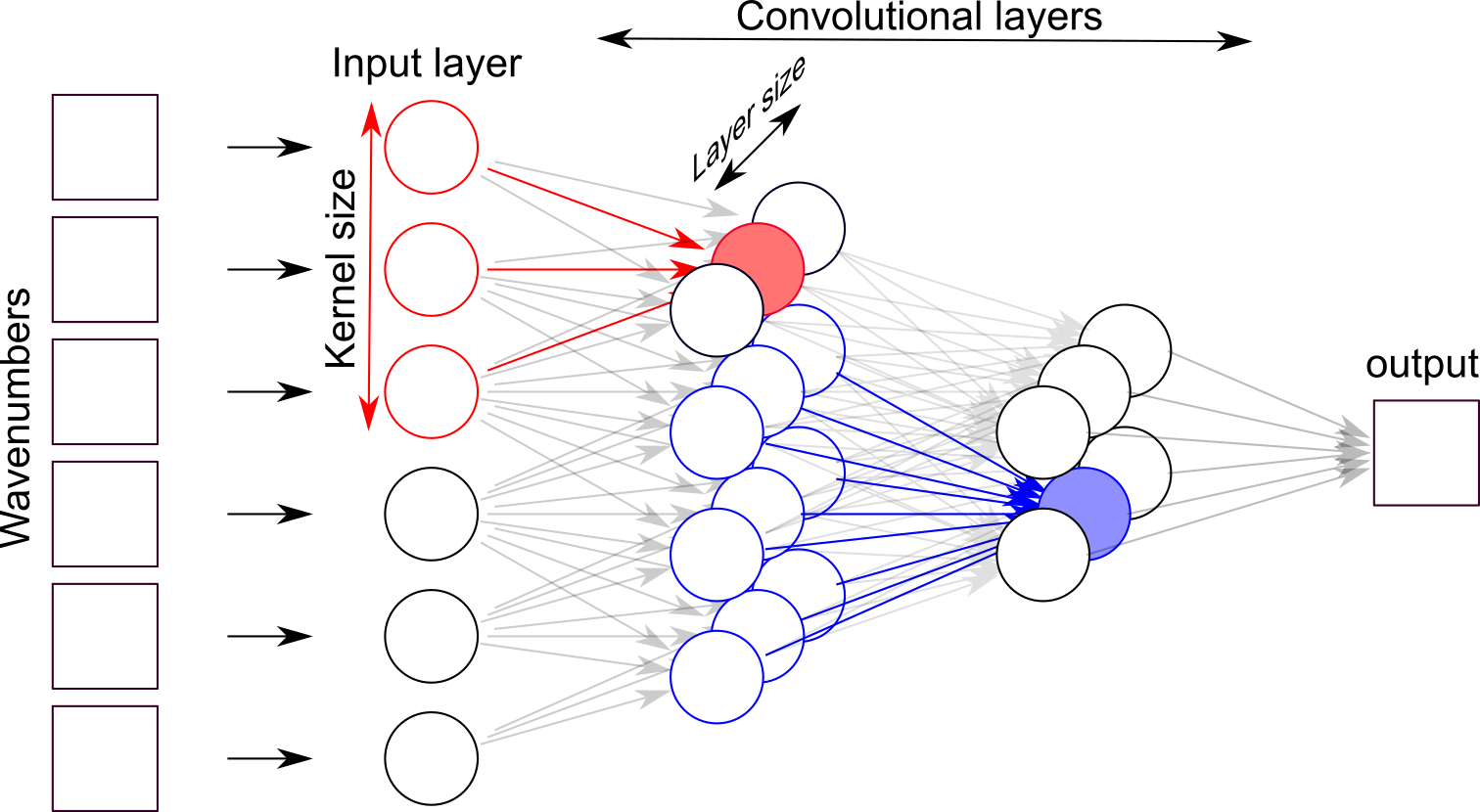}
 \caption{Example of a CNN, where kernel and layer sizes are marked. A few neurons have been highlighted in order to show the flow of input and output. Figure reprinted from ref. \cite{Jernelv2020}.}
 \label{fgr:cnn}
\end{figure*}

In convolutional neural networks (CNNs) the hidden layers convolve the input and then give the result to the next layer, see Fig. \ref{fgr:cnn}. Consequently, the layers are not fully connected. For image analysis this makes the neural network less computationally expensive, and a deeper network can be used. For one-dimensional data computational power is a lesser concern, but it has been suggested that CNNs may avoid issues with overfitting previously seen in ANNs. The relatively few samples in spectroscopy datasets and large amounts of features makes ANNs very prone to overfitting. 

In a CNN, data is divided into one or several kernels, which will be associated with different sections of the input data \cite{Goodfellow2016}. The convolutional layer works by convolving, or shifting, the kernel from the first to the last variable. The kernel thereby acts as a filter, and can be shifted by different lengths called \textit{strides}. 

We tested both the Adam optimiser and the stochastic gradient descent (SGD) optimiser to train the neural network. These optimisers have a \textit{learning rate} parameter, which indicates how much the weights are changed during training. SGD has the additional \textit{momentum} parameter, which takes into account previous weight updates in order to converge faster. We also tested a \textit{dropout} rate, which is another method used to handle overfitting by setting the output of some randomly selected neurons to zero.

The model parameters used for the CNN in this work are summarised in Table \ref{tbl:CNN}, with the tested parameter ranges and the step sizes. The Keras framework was used to implement the CNNs in this study \cite{Ketkar2017}.
{
\renewcommand{\arraystretch}{1.5}
\begin{table}[h]
\centering
  \caption{Model parameters in the CNN, with the parameter ranges investigated in this study and the step sizes used}
  \label{tbl:CNN}
  \begin{tabular*}{0.48\textwidth}{@{\extracolsep{\fill}}lll}
    \hline
    Parameter & Range & Step  \\
    \hline
    Kernel size & 5--90 & 1 \\
   	Layer size & 5--50 & 1 \\
   	Strides & 1--25 & 1\\
   	SGD momentum & 0.2--0.9 & 0.1 \\
   Learning rate & 10\textsuperscript{-6}--10\textsuperscript{-2}  &  factor of 10\\
   Dropout rate & 0--0.3  & 0.1 \\
  \end{tabular*}
\end{table}
}

\subsubsection{Regression methods for comparison}

\emph{Principal component regression:}
Principal component regression (PCR) is based on principal component analysis (PCA), where the original data is projected onto a smaller variable subspace with linearly uncorrelated variables, which are called principal components (PCs) \cite{Naas1988}. The number of PCs used in each model was chosen through cross-validation, from the range 3--25 PCs. 

\emph{Partial least-squares regression:}
Partial least-squares regression (PLSR) is one of the most commonly used multivariate analysis methods in spectroscopy \cite{Wold2001}. PLSR finds a regression model by projecting both the dependent and independent variables to a new space defined by a set of latent variables. The number of PLSR latent variables (LVs) was chosen through cross-validation on training data, from the range 3--20 LVs.

\emph{Random forest regression:}
Random forest (RF) is an ensemble learning method that can be used for several purposes, and has been used successfully for regression of spectral data \cite{Liaw2002}. Random forest regression works by creating multiple decision trees, and combining these for regression analysis. RF regression was evaluated with maximum tree depth for each dataset and 200--600 decision trees.

\emph{Elastic net:}
Elastic net (Net) is another ensemble learning method that combines the penalty variables from lasso and ridge regression ($\mathbf{L}_1$ and $\mathbf{L}_2$, respectively) \cite{Zou2005}. Elastic net was evaluated with an $\mathbf{L}_1/\mathbf{L}_2$ ratio of 0--1 in increments of 0.1.

\subsubsection{Classification methods for comparison}

\emph{Partials least-squares discriminant analysis:}
PLS discriminant analysis (PLS-DA) uses classical PLSR on categorical variables, which is enabled by dividing the categorical variable into dummy variables that describe the categories (also called one hot encoding) \cite{Brereton2014}. Optimal PLS-DA models were chosen based on LVs in the range 3--20.

\emph{$k$-nearest neighbours:}
k-nearest neighbours ($k$NN) is a simple learning method that classifies predicted data points based on their distance from the $k$ nearest neighbours in the training set \cite{Altman1992}. The distance between test data and training data can be calculated in different ways, e.g. Euclidean, Chebyshev, or cosine. This study used the Euclidean distance method as it is most common, and the models have been optimised for $k$ in the range 3--10.

\emph{Support vector machines:}
Support vector machines (SVMs) are a class of learning methods that try to find hyperplanes in a multidimensional space in order to divide data points into separate classes \cite{Cortes1995}. A support vector classifier with a linear kernel was used for this study.

\emph{Logistic regression:}
Logistic regression (LogReg) is a categorical method used to fit data with a special logistic function \cite{Hastie1987}. LogReg models can be considered simplified ANNs with no hidden layers. We used a LogReg model with an  $\mathbf{L}_2$ penalty.

\subsubsection{Pre-processing methods}
Various methods can be used on spectroscopic data to correct for scattering effects, including standard normal variate (SNV) and multiplicative scattering correction (MSC). Spectral derivatives can also widely used to remove unwanted spectral effects. 

Filtering or smoothing methods can be used alone on data to reduce noise. In addition, filtering methods are used to improve the noise in spectral derivatives. Savitzky-Golay (SG) differentiation is by far the most common method used. In the smoothing step for SG filtering a polynomial is fitted to the data points in the window using linear least-squares. 

The pre-processing methods used in this study can be divided into five separate steps, where some have different possible methods and associated variable parameters:

\begin{itemize}
\item\emph{Data binning:} Binning together 1--16 data points
\item\emph{Scatter correction:}
	\begin{itemize}
		\item Normalisation
		\item Standard normal variate (SNV)
		\item Multiplicative scattering correction (MSC)
	\end{itemize}
\item\emph{Smoothing/filtering:} 
	\begin{itemize}
		\item Savitzky-Golay filter (order 1--3, width 5--21)
		\item Fourier filter
		\item Finite/infinite impulse response filters (Butterworth, Hamming, moving average)
	\end{itemize}
\item\emph{Baseline corrections:}
	\begin{itemize}
		\item Subtract constant value
		\item Subtract linear background
		\item Spectral derivative (1st or 2nd derivative)
	\end{itemize}
\item\emph{Scaling:}
	\begin{itemize}
		\item Mean centering
		\item Scaling
	\end{itemize}
\end{itemize}

Note that this is a small subset of all possible pre-processing methods that exist, and we chose some general methods that are often applied to spectroscopy data \cite{Rinnan2009, Engel2013}. There is a lack of consensus regarding the best order of different pre-processing steps, and recent studies have shown that there is no clear optimal order \cite{Butler2018}, although the order can influence the prediction accuracy.

\subsubsection{Feature selection}
\begin{figure}[h]
\centering
  \includegraphics[height=7.8cm]{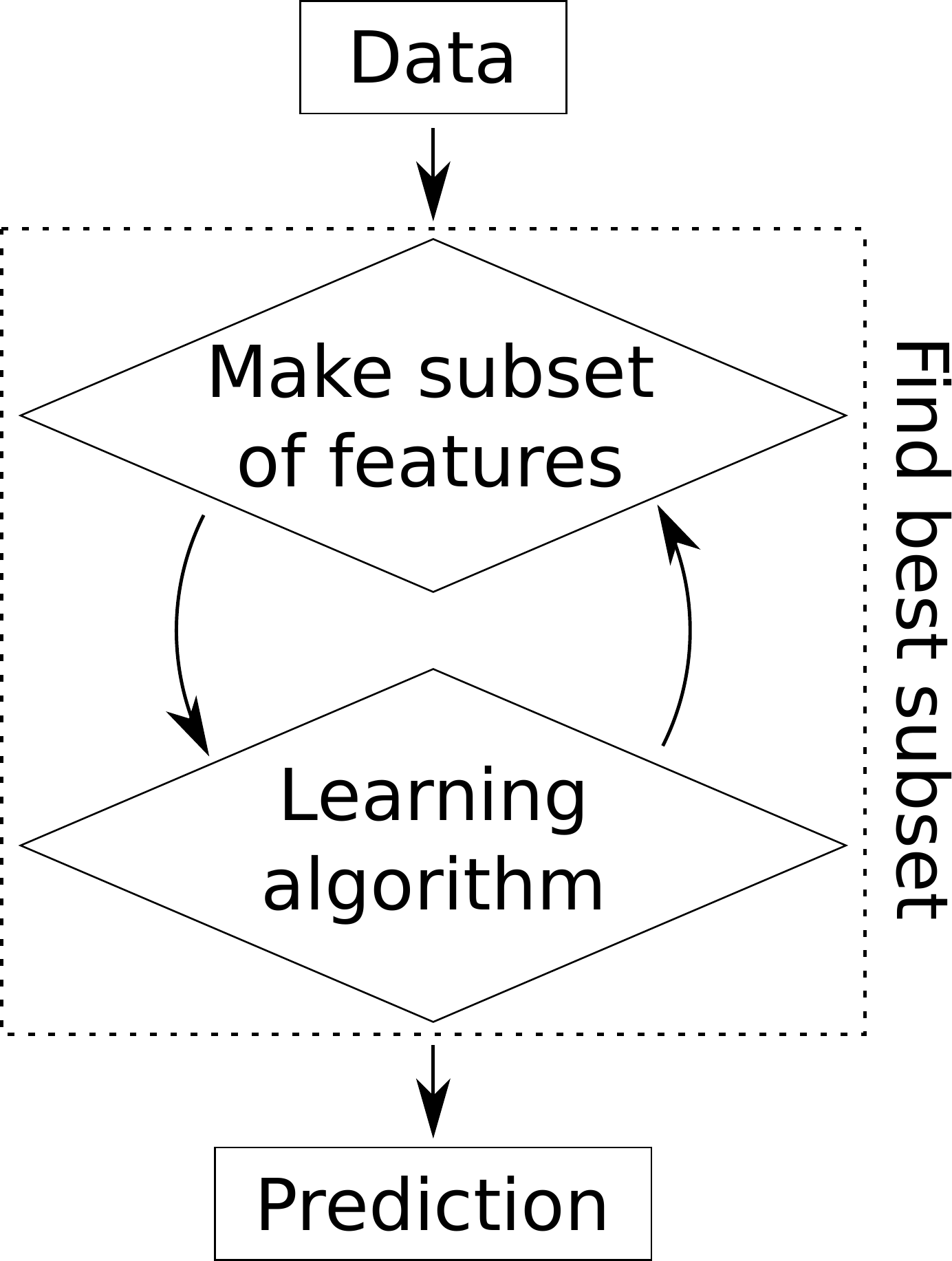}
  \caption{General work flow of wrapper methods for feature selection.}
  \label{fgr:flowchart}
\end{figure}

Spectral data can contain many hundreds or thousands of spectral features. For most types of data, the spectra contain features that are either irrelevant or redundant. Many types of feature selection methods have been developed, see refs. \cite{Saeys2007, Mehmood2012} for extensive reviews. We tested three different wrapper methods for the datasets in this study, which are briefly described below.  Wrapper methods generally work by searching through several subsets of features and choosing a feature subset based on prediction accuracy, see Fig. \ref{fgr:flowchart} for a schematic.

\emph{Sequential selection:} In forward selection, one starts by choosing the wavenumber which has the highest correlation with the Y-variable. The wavenumbers that give the highest model improvements are then added iteratively, until no improvements occur when more features are added. Backward elimination, on the other hand, starts with all the features, and tries to eliminate features that worsen the prediction. Sequential, or stepwise, selection combines these two methods by reconsidering variables for re-introduction or removal at each step.

\emph{Moving Window:}
A window is moved over the spectrum, and regression models are built using the data in the window position \cite{Jiang2002}. The window size can be varied over a set range. Optimal spectral intervals can then be chosen, based on the prediction error and number of regression components used in the model. Some moving window algorithms also try to optimise the selection of multiple windows \cite{Kasemsumran2004}. 

\emph{Genetic Algorithm:} Genetic algorithms (GA) for wavelength selection exist in many variants \cite{Jouan-Rimbaud1995}. Generally, a population consisting of $K$ vectors with length $n$ are initialised. Two vectors are then selected from the population as parent vectors, and these are subjected to crossover and/or mutation, resulting in offspring vectors. The offspring are then evaluated based on their fitness, where they are either rejected or they replace the worst member of the population. Regression models are then built using the resulting vectors after a set amount of iterations.

\subsection{Example datasets}

Datasets of different sample types acquired with different measurement methods have been analysed. A dataset with aqueous solutions was measured in a custom-built ATR-FTIR setup for this study. Other openly available datasets of food items or pharmaceutical tablets measured with FTIR, NIR or Raman spectroscopy have also been investigated. An overview of the dataset properties is provided in Table~\ref{tbl:datsets}. For datasets with no preset test set, such as the Tablets data, we separated the data randomly into training and test sets based on a 67\%$/$33\% split. These datasets were chosen in order to represent different measurement methods and wavelength ranges within vibrational spectroscopy.

{
\renewcommand{\arraystretch}{1.7}
\begin{table*}
\centering
  \caption{\ Characteristics of the datasets used for classification and regression analysis}
  \label{tbl:datsets}
  \begin{tabular*}{0.99\textwidth}{@{\extracolsep{\fill}}lllllll}
    \hline
    Data & Method & Wavelength range &Calibration samples & Validation samples & Features & Classes  \\
    \hline
    Tablets & NIR & 10507--7400 cm\textsuperscript{-1} &211 & 99 & 404 & 4\\
    Tablets & Raman & 3600-200 cm\textsuperscript{-1} &  82 & 38 & 3402 & 4\\
    Fruits & FTIR &1802--899 cm\textsuperscript{-1} &666 & 317 & 235 & 2\\
    Wines & FTIR & 5011--929 cm\textsuperscript{-1} & 30 & 14 & 842 & 4\\
    Solutions & FTIR & 4000--500 cm\textsuperscript{-1} & 60 & 30 & 1814 & NA \\
  \end{tabular*}
\end{table*}
}

The following spectral datasets are investigated in this study:
\begin{itemize}
	\item Solutions dataset, with aqueous solutions containing glucose, albumin, lactate, and urea, measured with ATR-FTIR. This dataset was acquired by the authors, see the next section for further details (dataset available from https://github.com/jernelv/SpecAnalysis)
	\item Tablets dataset, where the samples are categorised into four different tablet types with different relative amounts of active materiel, measured with both NIR and Raman spectroscopy \cite{Dyrby2002}. Both datasets were used for classification, while the Tablets NIR dataset was also used for regression analysis since relative weight percentage was also available (dataset available from http://www.models.life.ku.dk/Tablets)
	\item Wines dataset, where the samples are categorised into wines with different origin countries, measured with FTIR spectroscopy \cite{Skov2008} (dataset available from http://www.models.life.ku.dk/Wine\_GCMS\_FTIR)
	\item Fruit pur\'{e}es datasets measured with FTIR, where the samples are categorised into either strawberry or other pur\'{e}es \cite{Holland1998} (dataset available from https://csr.quadram.ac.uk/example-datasets-for-download/) 
\end{itemize}

The Solutions dataset was used in the regression analysis for prediction of glucose and albumin concentrations, and the Tablets NIR dataset was used for prediction of the relative weight percentage of the active material. The Tablets NIR, Tablets Raman, Fruits, and Wines datasets were employed for classification tasks of the categories described above.

\subsubsection{Solutions dataset}
In total, 90 unique aqueous solutions were made with four analytes, with the concentration ranges shown in Table \ref{tbl:conc}. Samples were made by dissolving the analytes in a phosphate-buffered saline solution (PBS). PBS is a buffer solution that helps maintain the pH in the solutions, and was made by dissolving PBS powder (Wako) in demineralised water. Glucose, albumin, urea, and lactate were then added to the solutions in varying concentrations.

{
\renewcommand{\arraystretch}{1.7}
\begin{table}[h]
\small
\centering
  \caption{\ Concentration ranges for the sample analytes used in the Solutions dataset}
  \label{tbl:conc}
  \begin{tabular*}{0.48\textwidth}{@{\extracolsep{\fill}}ll}
    \hline
    Analyte & Concentration ranges [mg/dl]  \\
    \hline
    Glucose & 0--800 \\
    Lactate & 0--90 \\
    Albumin & 0--6000 \\
    Urea & 0--200 \\
  \end{tabular*}
\end{table}
}

The sample concentrations were determined with an optimal design model in order to randomise the concentrations while still filling the entire design space. The experimental design was made using a quadratic Scheffe model with A-optimality design.

60 samples were used for the training set and 30 samples for the prediction set. Samples were assigned randomly to the different sets.

Spectra were recorded using an FTIR spectrometer (Tensor27, Bruker, Germany). This spectrometer was modified in a custom setup for ATR measurements with multi-reflection prisms. Guiding optics and a hollow-core fiber was used to couple the light into a ZnS ATR crystal. This experimental setup has previously been used by the Matsuura group, see e.g. Kasahara et al. \cite{Kasahara2018} for more details. 

Data acquisition and initial spectral processing was performed using the OPUS software package (ver 6.0, Bruker Optics, Germany). 32 interferograms were co-added for each measurement, and scans were performed with a nominal resolution of 4 cm\textsuperscript{-1}. A zero-filling factor of 2 was used before the Fourier transform, which reduced the datapoint spacing to approximately 2 cm\textsuperscript{-1}, together with a 3-term Blackman-Harris apodisation. 

\begin{figure}[h]
\centering
  \includegraphics[width=0.49\textwidth]{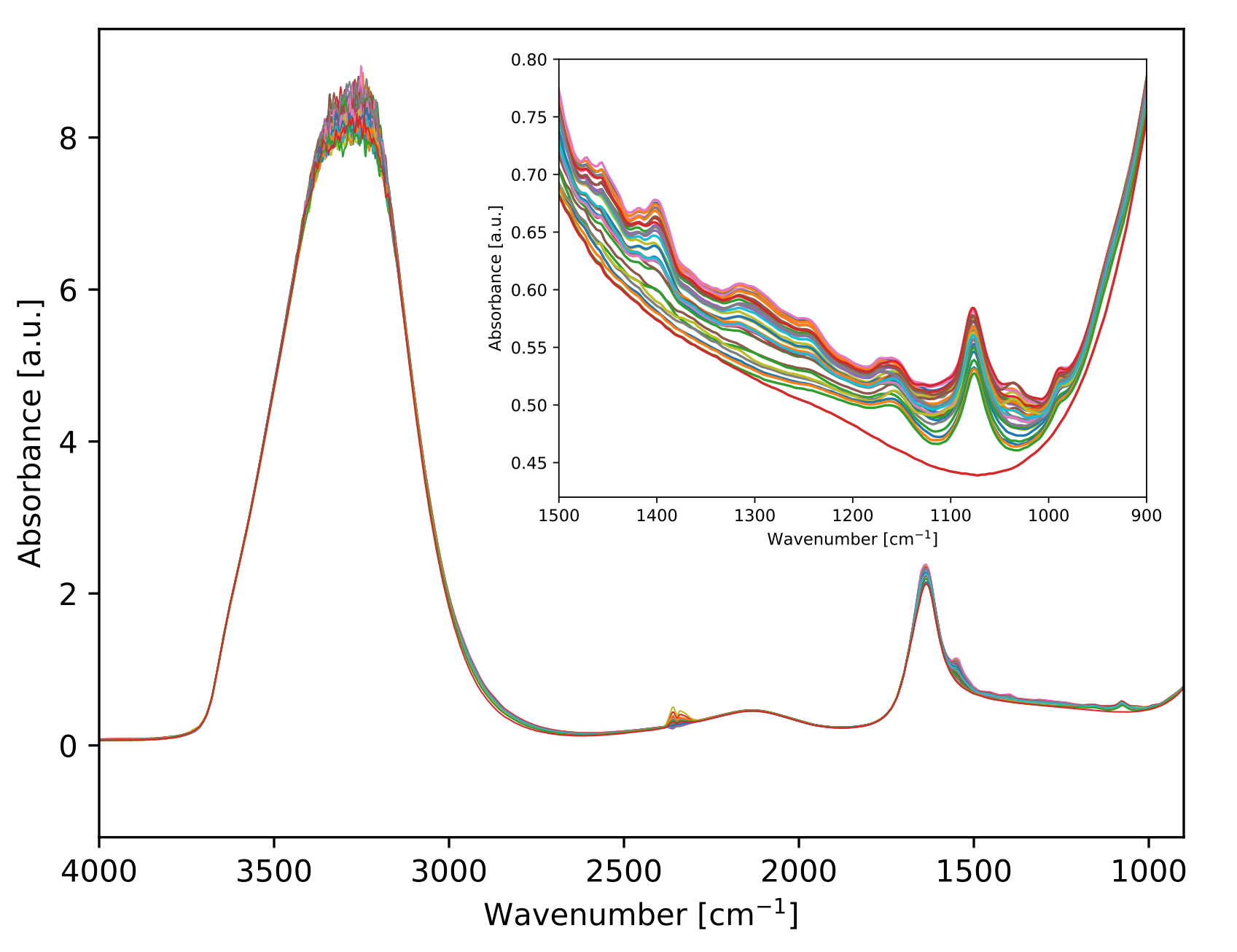}
  \caption{Plot of the solutions spectra acquired in FTIR. The inset shows the fingerprint region.}
  \label{fgr:ftir}
\end{figure}

Plots of the acquired spectra are shown in Fig.~\ref{fgr:ftir} for the range 4000-900~cm\textsuperscript{-1}. The inset shows the range 1500-900~cm\textsuperscript{-1}, which has the most informative spectral bands for the analytes in this dataset. This data was cut down to the 3000--800~cm\textsuperscript{-1} range for the analysis, due to large amounts of noise in the high absorption water bands.

The samples were placed manually on the ATR crystal (approx. 0.25 mL) for measurements. Each sample was measured 5 times in order to test measurement repeatability, with at least 2 repeat measurement done on the same day to check within-day variation. Background measurements of the ambient background were done every 30 minutes. The measurement series were performed over a period of two months.

\subsection{Model evaluation}
CNN was compared to other regression and classification models by applying the methods to the datasets in four different cases: raw data (i.e. no pre-processing), pre-processing prior to modelling, feature selection prior to modelling, and finally both pre-processing and feature selection prior to modelling. For CNN we mainly looked at raw data and pre-processing methods, which was in part due to the long processing time required for CNN compared to the other methods. Important spectral regions can be identified with neural networks through for example stability feature selection or network pruning, and some examples of this are included in the Supplementary Information.

The best combination of pre-processing methods is not necessarily the same for different classification or regression models. To ensure that the best methods were used for each model, we therefore tested all possible combinations of the pre-processing methods mentioned above. The method combination with the highest accuracy was then chosen based on leave-one-out cross-validation (LOOCV) on the training set. 

Similarly, the optimal feature subset might not be the same for all models. We therefore did feature selection separately all methods and all datasets, and the best features selection was chosen based on LOOCV on the training dataset. All three wrapper methods described above were tested.

For the classification datasets, the models were evaluated based on the percentage of correctly classified samples in the test sets. For the regression analysis, the models were evaluated based on the coefficient of determination (R$^2$). The coefficient of determination indicates how close the predicted data correlate with the regression line. The root-mean-square error of prediction (RMSEP), which is a scale-dependent error value, was also calculated and is reported in the Supplementary Information. The best model parameters were chosen based on LOOCV on the training data in all cases, and prediction accuracy presented later was then calculated from the validation data.

\section{Results}

\subsection{Regression}
\begin{figure*}
\centering
  \includegraphics[height=7.2cm]{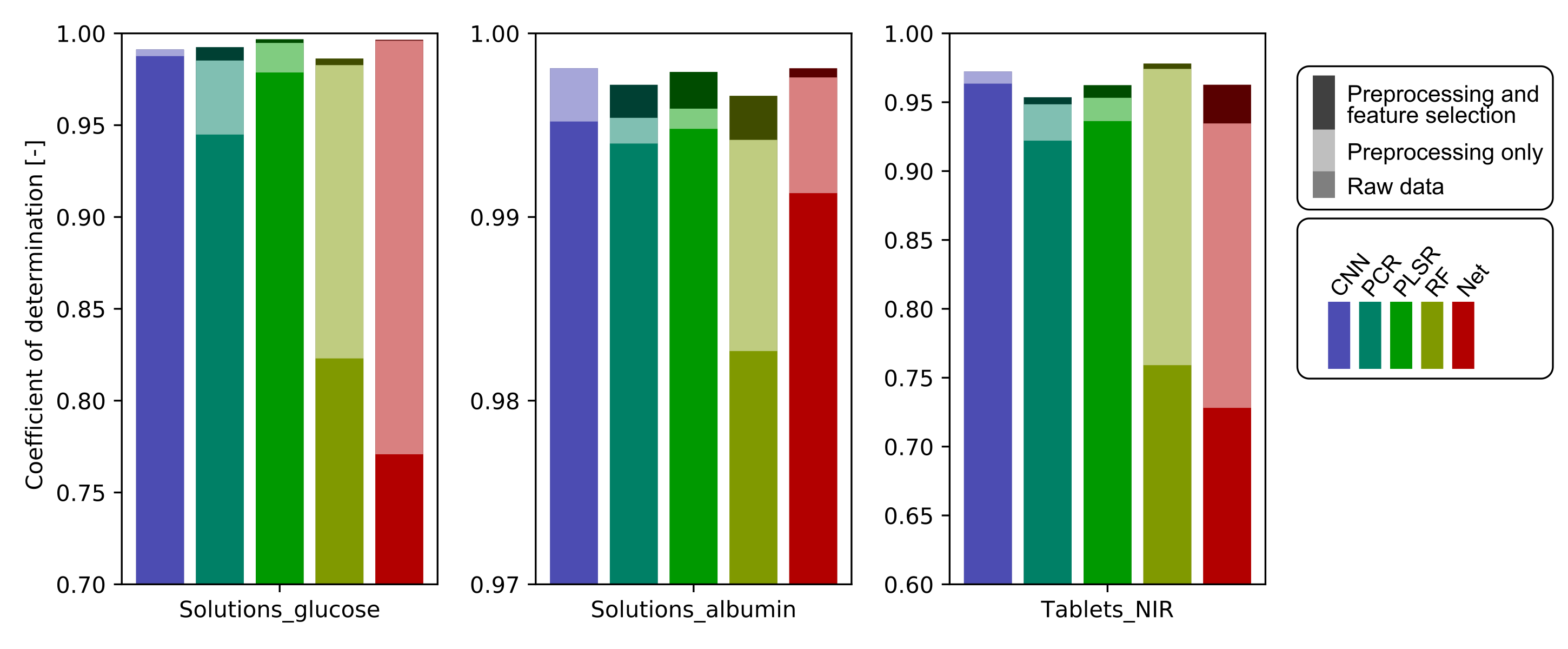}
  \caption{Plot of coefficient of determination for the regression analysis. This is shown for raw data, pre-processed data, and use of both pre-processing and feature selection. Note the different scales for each subplot. Results for feature selection alone can also be found in the Supplementary Information.}
  \label{fgr:regression}
\end{figure*}

The results of the regression analysis are summarised in Fig. \ref{fgr:regression}, with the coefficient of determination (R$^2$) of the regression methods.  Results are shown for regression on raw data, regression after pre-processing, and regression after pre-processing and feature selection. Full tables including the results for regression with only feature selection are included in the Supplementary Information. 

For the raw data (no pre-processing) our CNN model outperformed all the other regression methods for all datasets. PLSR and PCR, which are commonly used chemometric methods, generally performed well (R\textsuperscript{2}$>$0.92) with PLSR being somewhat better than PCR. Random forest and elastic net had much worse prediction accuracy than the other methods for raw data, but had comparable performance after pre-processing or feature selection. 

The CNN model also improved with application of pre-processing methods. After use of both pre-processing and feature selection on the other regression models, the CNN no longer had the singularly best performance on any of the datasets. However, the prediction accuracy was good for all the models tested here, with R\textsuperscript{2}$>$0.98 and R\textsuperscript{2}$>$0.99 for glucose and albumin in the Solutions dataset, respectively, and R\textsuperscript{2}$>$0.95 in the Tablets NIR dataset. 

\subsection{Classification}
\begin{figure*}
\centering
  \includegraphics[height=7.8cm]{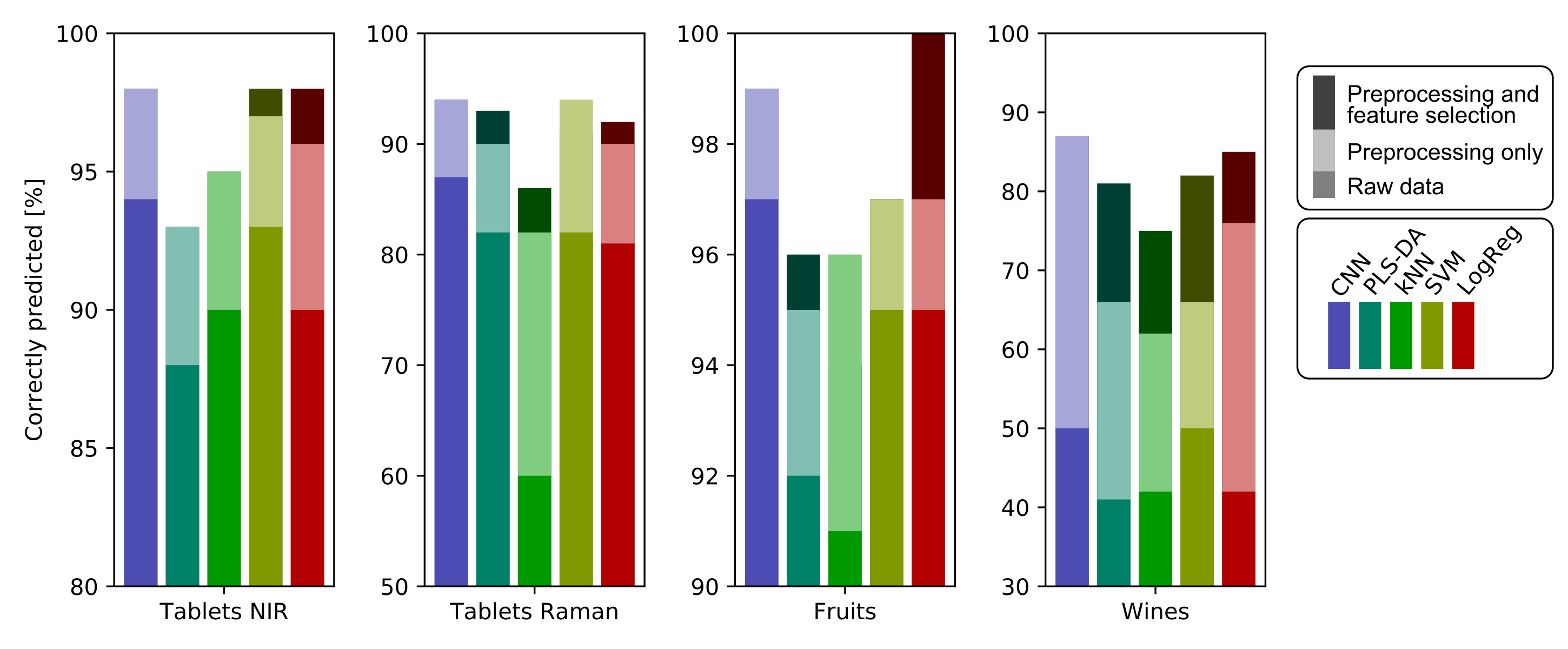}
  \caption{Plot of percentage correctly classified samples in the classification datasets. This is shown for raw data, pre-processed data, and use of both pre-processing and feature selection. Note the different scales for each subplot. Results for feature selection alone can also be found in the Supplementary Information.}
  \label{fgr:classify}
\end{figure*}

The results of the classification study are summarised in Fig. \ref{fgr:classify}, with the percentage accuracy of each classification method. Results are shown for the same cases as in Fig. \ref{fgr:regression}, namely raw data, pre-processing, and pre-processing together with feature selection. For the case where only feature selection was applied to the data the results can be found in the Supplementary Information.

For the raw data (no pre-processing), the CNN outperformed PLS-DA and $k$NN for all datasets. However, the percentage point difference between CNN and PLS-DA was not very large ($<$10\%). The difference between CNN and $k$NN was also generally small, but for the Tablets Raman dataset kNN performed much worse than all other methods. The performance of SVM and LogReg were generally comparable to CNN.

For data with either pre-processing or feature selection, there was improvement for nearly all the methods. CNN also experienced improvement with pre-processing. The improvement was most significant for the Wines dataset, with an increase of more than 20 percentage points for all methods, as the classification accuracy was lowest for the raw data from this dataset.

Using both pre-processing and feature selection on the datasets gave the largest improvement for many methods. This also made the performance of CNN, SVM, and LogReg more similar. In some cases, especially for the $k$NN, feature selection in addition to other pre-processing methods did not lead to an increase in the prediction accuracy.

\subsection{CNN parameters}
Parameters for the CNN model were fine-tuned for each dataset, and the chosen parameters are summarised in the Supplementary Information. Some parameters, such as optimal kernel size, varied a lot between different datasets. 

Two common methods for weight adjustment were used for this study, the Adam and the SGD optimisers. The SGD optimiser converged faster than the Adam optimiser, but did not always find an equally good prediction. 

We also tested the number of layers used for the CNN model, up to 4 layers. The performance declined when using more than 2 layers for all datasets. This demonstrates that the CNN is prone to overfitting with the use of more layers.

\begin{figure}[h]
\centering
  \includegraphics[height=5.6cm]{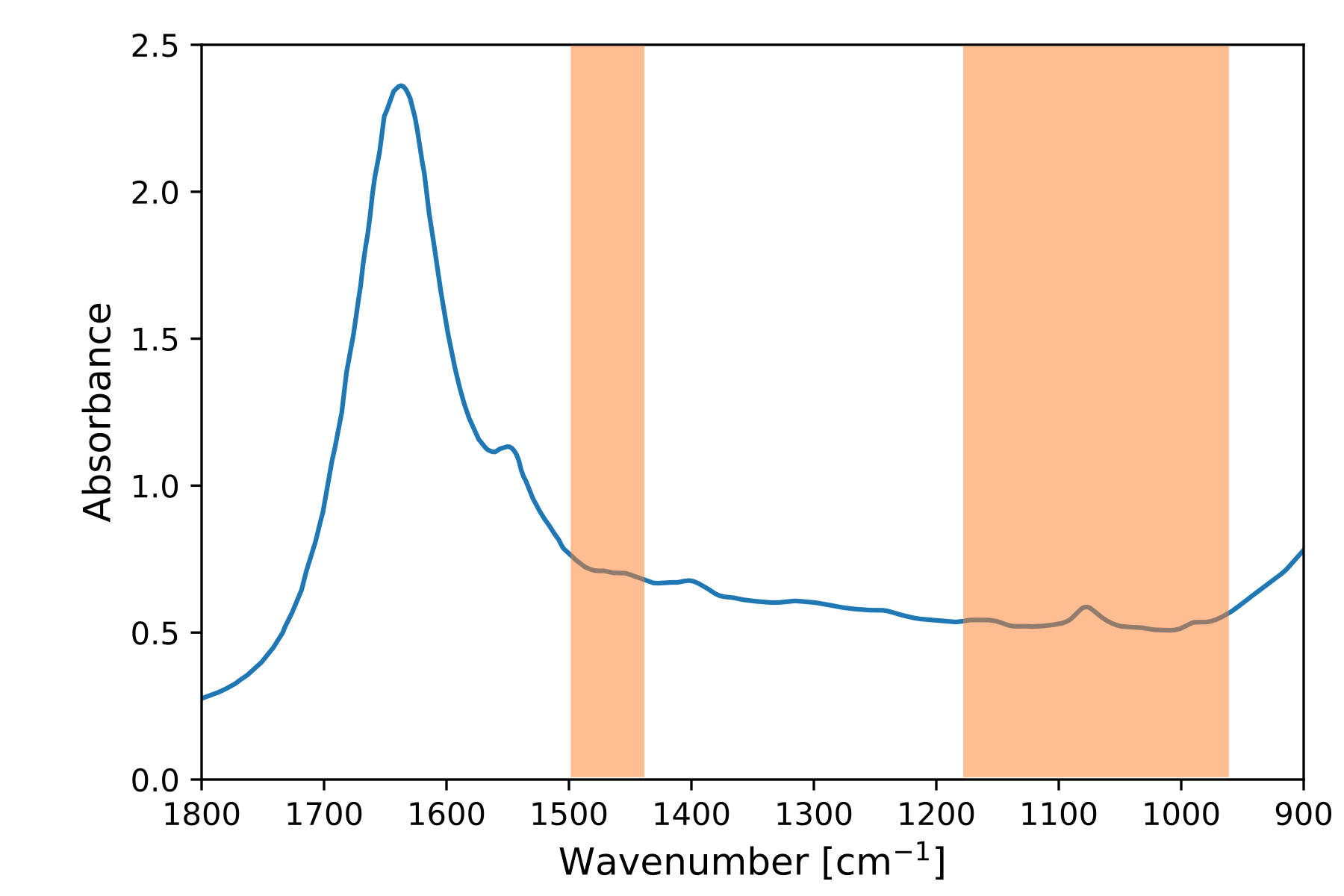}
  \caption{Important spectral regions in the analysis of glucose concentrations in the Solutions dataset, plotted for the 1800--900 cm\textsuperscript{-1} region. The spectral regions were determined by stability feature selection based on the CNN model.}
  \label{fgr:glucose_importantregions}
\end{figure}

Important spectral regions may be identified with CNNs using for example stability feature selection or network pruning. We performed a preliminary investigation of identification of important spectral regions in the datasets used for regression analysis. One example is shown in Fig. \ref{fgr:glucose_importantregions}, where regions relevant to glucose levels in the Solutions dataset have been identified. Note that only the fingerprint region 1800--900 cm\textsuperscript{-1} is shown here, as no important spectral regions were identified outside of this area. The stability feature selection recognised the spectral regions at approximately 1180-970 cm\textsuperscript{-1} and 1500-1430 cm\textsuperscript{-1}, which corresponds well with glucose absorption bands. There was minimal change in identification of spectral regions between raw data and pre-processed data, which demonstrates that the CNN model is able to consistently identify relevant regions. More examples of the stability feature selection are shown in the Supplementary Information.

\subsection{Use of pre-processing and feature selection}
Pre-processing improved the model accuracy for all models, both for the classification and regression analysis. The specific pre-processing methods applied to each dataset and each model are summarised in the Supplementary Information. For most classification models, at least noise filtering and some type of scaling were used in the best pre-processing strategy. CNN and logistic regression were particularly sensitive to scaling, as expected. For the Wines, Fruits, and Solutions datasets, spectral differentiation and filtering were also useful. Spectral differentiation has the benefit of removing baseline effects, as well as linear trends for the second derivative. Filtering prior to spectral differentiation also avoids the large amplification in noise that can otherwise occur. A small binning factor of 2 or 4 was also found to improve outcomes in several cases. Binning can improve SNR and leads to a dimension reduction; two factors that may both be beneficial for several machine learning methods. However, larger binning factors were mostly not positive, and may have obscured spectral information. Within each dataset, similar pre-processing steps were found to be optimal for the different modelling methods, indicating that pre-processing was mainly dataset-dependent.

Feature selection, either alone or together with pre-processing methods, often improved the prediction accuracy for both regression analysis and classification. Of the three wrapper methods that were tested, genetic algorithm resulted in the best performance, and generally selected similar relevant areas as in the stability feature selection for the CNN. Sequential forward selection and moving window did not always lead to improved accuracy, and particularly sequential forward selection seemed to get stuck in local maxima. Feature selection also improved the performance of the embedded methods with built-in feature selection, such as elastic net. Although elastic net is an embedded method, it is known to perform worse if there are many more features than observations, and often started off with low prediction accuracy on the raw data.

\subsection{Discussion}
CNN models can outperform standard chemometric methods, both for classification and regression analysis. CNN generally has a somewhat better performance than other methods when raw data is analysed directly. However, this advantage becomes much smaller when pre-processing and/or feature selection methods are applied to the datasets. 

CNN has been suggested as a classification method that would depend less on pre-processing as compared to standard chemometric methods, e.g. by Acquarelli et al. \citep{Acquarelli2017}. In theory, CNN could therefore be used directly without the need for a time-consuming search of the optimal pre-processing steps and methods. However, this ignores the need for parameter tuning inherent in the CNN. A CNN includes several parameters such as layer size, kernel size, stride, etc., and the parameter choice can vary significantly between datasets. Hence, using CNN to avoid a choice of pre-processing methods may ultimately not save any computational time. Methods such as SVM and PLSR generally have fewer parameters, and are therefore easier to tune to optimal conditions, with only a small loss in accuracy as compared to CNN.

\section{Conclusion}
In this study, a convolutional neural network (CNN) was applied for classification and regression analysis of various datasets acquired with spectroscopic methods. We compared the CNN to several standard methods used for classification and regression in chemometrics. The CNN generally performed better on raw data compared to both the classification and regression models. This was also the case when either pre-processing or feature selection were applied to the datasets, although the advantage of CNN decreased. In general, application of appropriate pre-processing and feature selection significantly improves the accuracy of most models.  

\section*{Conflicts of interest}
There are no conflicts of interest to declare.

\section*{Acknowledgments}
This work was supported by the Research Council of Norway through the Double Intraperitoneal Artificial Pancreas project, grant number 248872. The project is part of Centre for Digital Life Norway and is also supported by the Research Council of Norway's grant 248810.

The authors would like to thank Dr. Saiko Kino for training on the FTIR spectrometer setup at Tohoku University.

%\balance


%%%REFERENCES%%%
\bibliography{analysis_ref} %You need to replace "rsc" on this line with the name of your .bib file
\bibliographystyle{rsc} %the RSC's .bst file

\end{document}

\preprint{APS/123-QED}

\title{Supplementary Information for "Convolutional neural networks for classification and
regression analysis of one-dimensional spectral data"}% Force line breaks with \\

\author{Ine L. Jernelv}
 \email{ine.jernelv@ntnu.no}
\author{Dag Roar Hjelme}%
\author{Astrid Aksnes}
\affiliation{%
 Department of Electronics, Norwegian University of Science and Technology (NTNU), O.S. Bragstads plass 2A, 7491 Trondheim, Norway
}

\author{Yuji Matsuura}
\affiliation{
Graduate School of Biomedical Engineering, Tohoku University, Sendai, 980-8579, Japan
}%
\date{\today}

\begin{abstract}
This supplemental information contains results and figures that were not included in the main article for conciseness, or due to a lack of space.

\end{abstract}

\let\mytemp\clearpage
\let\clearpage\ 
\maketitle
\let\clearpage\mytemp

All analysis performed in this work was done in the software package SpecAnalysis, which has been made available online (https://github.com/jernelv/SpecAnalysis).

\section{CNN Parameters}

The parameters used for our CNN for the regression and classification datasets are shown in Table \ref{tbl:cnn_reg} and Table \ref{tbl:cnn_class}, respectively. Both the Adam and the SGD optimisers were tested, and the momentum parameter is given for the cases were the SGD optimiser had the best performance.

{
\renewcommand{\arraystretch}{2}
\begin{table}[h]
\centering
  \caption{CNN parameters used for the regression datasets.}
  \label{tbl:cnn_reg}
  \begin{tabular*}{0.9\textwidth}{@{\extracolsep{\fill}}lcccccc}
    \hline
    Dataset & Layer size & Kernel size & Learning rate & Stride & Momentum & Dropout \\
    \hline
    Solutions glucose & 10 & 10 & 10\textsuperscript{-3} & 4 & 0.9& 0\\
    Solutions albumin & 19 & 16 & 10\textsuperscript{-3} & 5 & -- & 0\\
    Tablets NIR & 25 & 45 & 10\textsuperscript{-3} & 3 & 0.8 & 0.1\\
    \hline
  \end{tabular*}
\end{table}
}

{
\renewcommand{\arraystretch}{2}
\begin{table}[h]
\centering
  \caption{CNN parameters used for the classification datasets.}
  \label{tbl:cnn_class}
  \begin{tabular*}{0.9\textwidth}{@{\extracolsep{\fill}}lcccccc}
    \hline
    Dataset & Layer size & Kernel size & Learning rate & Stride & Momentum & Dropout \\
    \hline
    Tablets NIR & 6 & 5 & 10\textsuperscript{-3} & 26 & -- & 0\\
    Tablets Raman & 12 & 37 & 10\textsuperscript{-3} & 6 & 0.9 & 0\\
    Fruits & 8 & 52 & 10\textsuperscript{-3} &12 & -- & 0 \\
    Wines & 4 & 46 & 10\textsuperscript{-3} & 6 & 0.8 & 0 \\
    \hline
  \end{tabular*}
\end{table}
}

\newpage
\section{Prediction Accuracy for Classification Methods}
Table \ref{tbl:classify} shows the prediction accuracy for the classification methods in four different experiments: classification on raw data, classification on pre-processed data, classification on data after feature selection, and classification after both pre-processing and feature selection. Note that feature selection was not done on CNN.

{
\renewcommand{\arraystretch}{1.8}
\begin{table}[h]
  \caption{\ Classification accuracy [\%] on data, with and without pre-processing and feature selection. The best accuracy for each model type and dataset is marked in bold.}
  \label{tbl:classify}
\begin{tabular*}{\textwidth}{@{\extracolsep{\fill}}lccccccccccc}
\hline
              & \multicolumn{5}{l}{No feature selection} &      &         &      &      &         \\
              \hline
              & \multicolumn{5}{l}{Raw data}             & \multicolumn{5}{l}{Pre-processed data} \\
              \hline
Dataset       & CNN   & PLS-DA   & kNN  & SVM  & LogReg  & CNN  & PLS-DA  & kNN  & SVM  & LogReg  \\
	\hline
Tablets NIR   &  \textbf{94}	& 88&	90	&93&	90&		\textbf{98}&	93	&95&	97&	96\\

Tablets Raman &  \textbf{87}&	82	&60	&82	&81	&	\textbf{94}	&90	&82	&\textbf{94}	&90\\

Fruits        &   \textbf{97}&	92	&91&	95&	95&		\textbf{99}&	95	&96	&97	&\textbf{99}\\

Wines         &   \textbf{50}&	41	&42&	\textbf{50}&	42	&	\textbf{87}	&66	&62	&66	&76\\

\hline
              &       &          &      &      &         &      &         &      &      &         \\
              \hline
              & \multicolumn{5}{l}{Feature selection}    &      &         &      &      &         \\
              \hline
              & \multicolumn{5}{l}{Raw data}             & \multicolumn{5}{l}{Pre-processed data} \\
              \hline
              &       &          &      &     &         &      &         &      &    &         \\
Tablets NIR   &   --   &   90     & 90    &   \textbf{95}   &   92      &     -- &    93     &  95    &  \textbf{98}      &   \textbf{98}      \\
Tablets Raman &   --  &    \textbf{88}      &  61    &   85   &     84    &  --   & 93        &  86 &   91   &    \textbf{92}     \\
Fruits        &   --    &     92     &  91    &   96   &     95    &   --   & 96        &  94    &  97    &   \textbf{100}      \\
Wines        &   --    &     58    &   45   &   60   &    62     &   --   &     81    &   75   &    82  &   85      \\
\hline
              &       &          &      &      &         &      &         &      &      &        
\end{tabular*}
\end{table}
}

\section{Prediction Accuracy for Regression Methods}
Table \ref{tbl:regress} shows the prediction accuracy for the regression in four different experiments: regression analysis on raw data, regression on pre-processed data, regression on data after feature selection, and regression after both pre-processing and feature selection.

{
\renewcommand{\arraystretch}{1.8}
\begin{table}
  \caption{\ Prediction accuracy with coefficient of determination (R$^2$), with and without pre-processing and feature selection. The best accuracy for each model type and dataset is marked in bold.}
  \label{tbl:regress}
\begin{tabular*}{\textwidth}{@{\extracolsep{\fill}}lccccccccccc}
\hline
              & \multicolumn{5}{l}{No feature selection} &      &         &      &      &         \\
              \hline
              & \multicolumn{5}{l}{Raw data}             & \multicolumn{5}{l}{Pre-processed data} \\
              \hline
Dataset       & CNN   & PCR   & PLSR  & RF & Net  & CNN  & PCR  & PLSR  & RF & Net  \\
	\hline
Glucose  &  \textbf{0.9876}	&0.9449	&0.9787&	0.823&	0.7708&	0.9912	&0.9852&	0.9948&	0.9827&	\textbf{0.996}\\
Albumin &  \textbf{0.9952}     &  0.994	&0.9948&	0.9827&	0.9913& \textbf{0.9981} &  0.9954	&0.9959&	0.9942&	0.9972\\
Tablets NIR        &  \textbf{0.9635} &0.9221&	0.9363	&0.7591 & 0.7281  & 0.9723 & 0.9485	&0.9532&	\textbf{0.9742}&	0.9364\\
\hline
              &       &          &      &      &         &      &         &      &      &         \\
              \hline
              & \multicolumn{5}{l}{Feature selection}    &      &         &      &      &         \\
              \hline
              & \multicolumn{5}{l}{Raw data}             & \multicolumn{5}{l}{Pre-processed data} \\
              \hline
              &       &          &      &     &         &      &         &      &    &         \\
Glucose   &   --   &  0.9771 & 0.9833 & 0.9792 & 0.9811     &     -- & 0.9925 & \textbf{0.9968} & 0.9863 & 0.9965     \\
Albumin &   --  &   0.9965 & 0.9979 & 0.9864 & 0.9957   &  --   & 0.9972 & 0.9979 & 0.9966 & \textbf{0.9981}     \\
Tablets NIR  &   --    &     0.9318 & 0.9422 & 0.8826 & 0.8115   &   --   &  0.9536 & 0.9624 & 0.9781 & 0.9627    \\
\hline
              &       &          &      &      &         &      &         &      &      &        
\end{tabular*}
\end{table}
}

{
\renewcommand{\arraystretch}{1.8}
\begin{table}
  \caption{\ Prediction accuracy with root-mean-square error of prediction (RMSEP), with and without pre-processing and feature selection. The best accuracy for each model type and dataset is marked in bold.}
  \label{tbl:regress2}
\begin{tabular*}{\textwidth}{@{\extracolsep{\fill}}lccccccccccc}
\hline
              & \multicolumn{5}{l}{No feature selection} &      &         &      &      &         \\
              \hline
              & \multicolumn{5}{l}{Raw data}             & \multicolumn{5}{l}{Pre-processed data} \\
              \hline
Dataset       & CNN   & PCR   & PLSR  & RF & Net  & CNN  & PCR  & PLSR  & RF & Net  \\
	\hline
Glucose  &  \textbf{26.58}&	57	&35.4&	102.1&	116.2	&20.4	&29.5	&17.5&	31.9&	\textbf{15.3}\\
Albumin &  \textbf{120}     &  135.6&	126.6&	230.7	&163.8   &   \textbf{75.4} & 119.5&	111.5&	133.5&	92.3   \\
Tablets NIR        &  \textbf{0.2653} & 0.3684&	0.3331	&0.6478 & 0.6733 & 0.2196&	0.2994	&0.2856&	\textbf{0.2132}&	0.333\\
\hline
              &       &          &      &      &         &      &         &      &      &         \\
              \hline
              & \multicolumn{5}{l}{Feature selection}    &      &         &      &      &         \\
              \hline
              & \multicolumn{5}{l}{Raw data}             & \multicolumn{5}{l}{Pre-processed data} \\
              \hline
              &       &          &      &     &         &      &         &      &    &         \\
Glucose   &   --   &  38.9 & 27.7& 33.2 & 29     &     -- & 19.4 & \textbf{15.1} & 26.5 & 15.2     \\
Albumin &   --  &   113.8 & 105.8 & 111.2 & 121.9   &  --   & 95.8 & 85.3 & 0.110 & 77    \\
Tablets NIR  &   --    &     0.3368 & 0.3173 & 0.4519 & 0.4806  &   --   &  0.2859 & 0.2656 & \textbf{0.2118} & 0.2628    \\
\hline
              &       &          &      &      &         &      &         &      &      &        
\end{tabular*}

\end{table}
}

\clearpage
\section{Pre-Processing Methods}
Several pre-processing methods were evaluated for this study, and an overview can be found in Table \ref{tbl:preprocess}. These pre-processing methods were divided into five separate steps, and all possible combinations were tested for each dataset and each model. A more in-depth description of the various methods can be found in the Readme for the SpecAnalysis software package (https://github.com/jernelv/SpecAnalysis) or in the references in the main article.
{
\renewcommand{\arraystretch}{1.8}
\begin{table}[h]
\centering
  \caption{\ Alternatives for pre-processing methods used in this study.}
  \label{tbl:preprocess}
  \begin{tabular*}{0.9\textwidth}{@{\extracolsep{\fill}}ll}
    \hline
    \textbf{1. Binning} \\
    \hline
    - Binning together 1, 2, 4, 8, or 16 datapoints  \\
    \hline
    \textbf{2. Scatter correction} \\
    \hline
   	- Normalisation \\
   	- Standard normal variate (SNV)\\
   	- Multiple scatter correction (MSC)\\
    \hline
    \textbf{3. Smoothing/filtering} \\
    \hline
   	- SG filter & Polynomial order: 1, 2, or 3 \\
   	& Filter width: 3, 5, ..., 21 data points\\
   	- Fourier filter & Window function: none, Blackman-Harris, Hamming, Hann\\
   	& Filter cutoff: 20, 21, ... 50 points in Fourier space\\
   	& Filter window size: 1.1, 1.2, 1.3\\
   	- Finite/infinite impulse response filters: & Butterworth, Hamming, moving average (MA)\\
    \hline
    \textbf{4. Baseline correction} \\
        \hline
   	- Subtract constant value \\
   	- Subtract linear background\\
   	- Spectral differentiation: 1st or 2nd derivative\\
    \hline
    \textbf{5. Scaling} \\
        \hline
   	- Mean centering \\
   	- Scaling\\
    \hline
  \end{tabular*}
\end{table}
}

\newpage

\section{Optimal Pre-Processing Methods for Classification}
The following tables show the optimal pre-processing methods applied to each dataset for the classification analysis.
{
\renewcommand{\arraystretch}{2}
\begin{table}[h]
\small
\centering
  \caption{\ Pre-processing methods applied to the Tablets NIR dataset.}
  \label{tbl:tablets_nir_preprocess}
  \begin{tabular*}{0.9\textwidth}{@{\extracolsep{\fill}}ll}
    \hline
    Classification & Pre-processing \\
    \hline
    CNN & Baseline correction, scaling  \\
    PLS-DA & Baseline correction \\
    kNN & Baseline correction and scaling\\
    SVM & Scaling \\
    LogReg & Binning, Butterworth filter, baseline correction, scaling \\
    \hline
  \end{tabular*}
\end{table}
}

{
\renewcommand{\arraystretch}{2}
\begin{table}[h]
\small
\centering
  \caption{\ Pre-processing methods applied to the Tablets Raman dataset.}
  \label{tbl:tablets_raman_preprocess}
  \begin{tabular*}{0.9\textwidth}{@{\extracolsep{\fill}}ll}
    \hline
    Classification & Pre-processing \\
    \hline
    CNN & Binning, baseline correction, scaling  \\
    PLS-DA & Binning, Fourier filter \\
    kNN & Binning, normalisation, SG filter, baseline correction, scaling\\
    SVM & Binning, SG filter, first derivative, scaling\\
    LogReg & Binning, MA filter, scaling \\
    \hline
  \end{tabular*}
\end{table}
}

{
\renewcommand{\arraystretch}{2}
\begin{table}[h]
\small
\centering
  \caption{\ Pre-processing methods applied to the Wines dataset.}
  \label{tbl:wines_preprocess}
  \begin{tabular*}{0.9\textwidth}{@{\extracolsep{\fill}}ll}
    \hline
    Classification & Pre-processing \\
    \hline
    CNN & Binning, SG filter, second derivative, scaling  \\
    PLS-DA & Binning, SG filter, second derivative \\
    kNN & Binning, MA filter, second derivative, scaling\\
    SVM & Binning, SG filter, second derivative, scaling\\
    LogReg & Binning, SG filter, second derivative, scaling \\
    \hline
  \end{tabular*}
\end{table}
}

{
\renewcommand{\arraystretch}{2}
\begin{table}[h]
\small
\centering
  \caption{\ Pre-processing methods applied to the Pur\'ees dataset.}
  \label{tbl:fruits_preprocess}
  \begin{tabular*}{0.9\textwidth}{@{\extracolsep{\fill}}ll}
    \hline
    Classification & Pre-processing \\
    \hline
    CNN & Binning, SG filter, second derivative, scaling  \\
    PLS-DA & Normalisation, first derivative, scaling \\
    kNN & First derivative, scaling\\
    SVM & Normalisation, scaling\\
    LogReg & Normalisation, subtract linear baseline, scaling \\
    \hline
  \end{tabular*}
\end{table}
}
\newpage
\section{Optimal Pre-Processing Methods for Regression}
The following tables show the sequency of the optimal pre-processing methods applied to each dataset for the regression analysis. 
{
\renewcommand{\arraystretch}{2}
\begin{table}[h]
\small
\centering
  \caption{\ Pre-processing methods applied to the Solutions dataset with glucose as the target analyte.}
  \label{tbl:glucose_preprocess}
  \begin{tabular*}{0.9\textwidth}{@{\extracolsep{\fill}}ll}
    \hline
    Regression & Pre-processing \\
    \hline
    CNN & Binning, MA filter, first derivative, scaling \\
    PCR & First derivative, scaling \\
    PLSR & SG filter, first derivative, scaling\\
    RF & Binning, SG filter, second derivative \\
    Net & SG filter, first derivative, scaling\\
    \hline
  \end{tabular*}
\end{table}
}

{
\renewcommand{\arraystretch}{2}
\begin{table}[h]
\small
\centering
  \caption{\ Pre-processing methods applied to the Solutions dataset with albumin as the target analyte.}
  \label{tbl:albumin_preprocess}
  \begin{tabular*}{0.9\textwidth}{@{\extracolsep{\fill}}ll}
    \hline
    Regression & Pre-processing \\
    \hline
    CNN & Binning, Butterworth filter, second derivative, scaling  \\
    PCR & Normalisation, SG filter, second derivative, scaling \\
    PLSR & Binning, SG filter, first derivative, scaling\\
    RF & Binning, SG filter, first derivative, scaling\\\
    Net & SG filter, scaling\\
    \hline
  \end{tabular*}
\end{table}
}

{
\renewcommand{\arraystretch}{2}
\begin{table}[h]
\small
\centering
  \caption{\ Pre-processing methods applied to the Tablets NIR dataset for regression analysis.}
  \label{tbl:tablets_nir_regress_preprocess}
  \begin{tabular*}{0.9\textwidth}{@{\extracolsep{\fill}}ll}
    \hline
    Regression & Pre-processing \\
    \hline
    CNN & Binning, MSC, MA filter, scaling  \\
    PCR & Binning, MSC, SG filter, subtracted linear baseline, scaling\\
    PLSR & Binning, MSC, SG filter\\
    RF & Binning, SNV, Butterworth filter, second derivative\\\
    Net & MSC, SG filter, first derivative, scaling\\
    \hline
  \end{tabular*}
\end{table}
}
\newpage
\section{Feature Selection with CNN}
In a preliminary analysis, stability feature selection was used on the regression data in order to identify the most important spectral regions. For the Solutions dataset, only the fingerprint region 1800--900 cm\textsuperscript{-1} has been plotted for clarity, as this was the area in the dataset where important information was identified by the CNN. Only minimal differences were seen in the important regions on the raw vs. the pre-processed data.

\begin{figure}[h]
\centering
  \includegraphics[height=7cm]{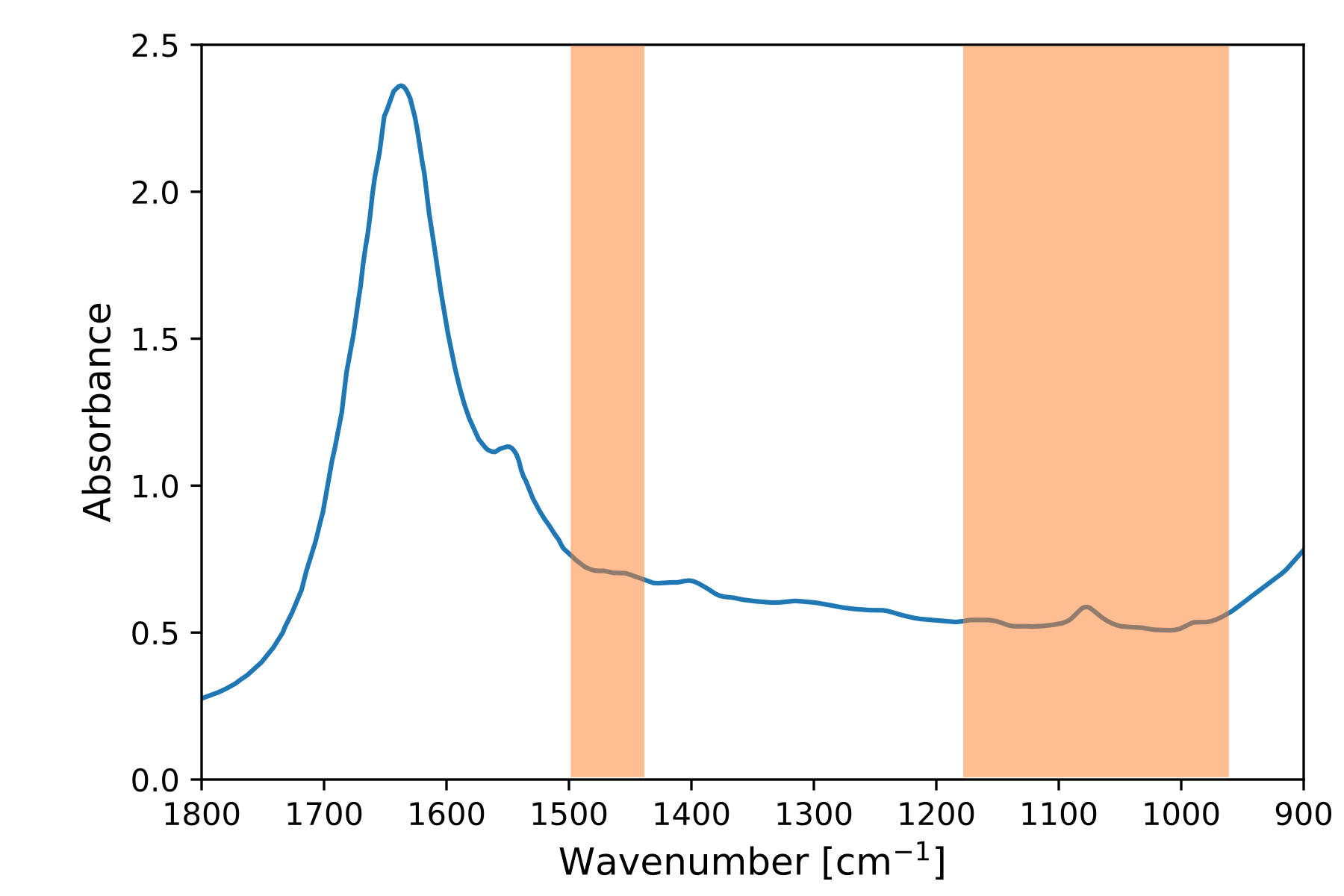}
  \caption{Important regions identified for CNN regression analysis of glucose in the Solutions dataset.}
  \label{fgr:glucose}
\end{figure}

\begin{figure}[h]
\centering
  \includegraphics[height=7cm]{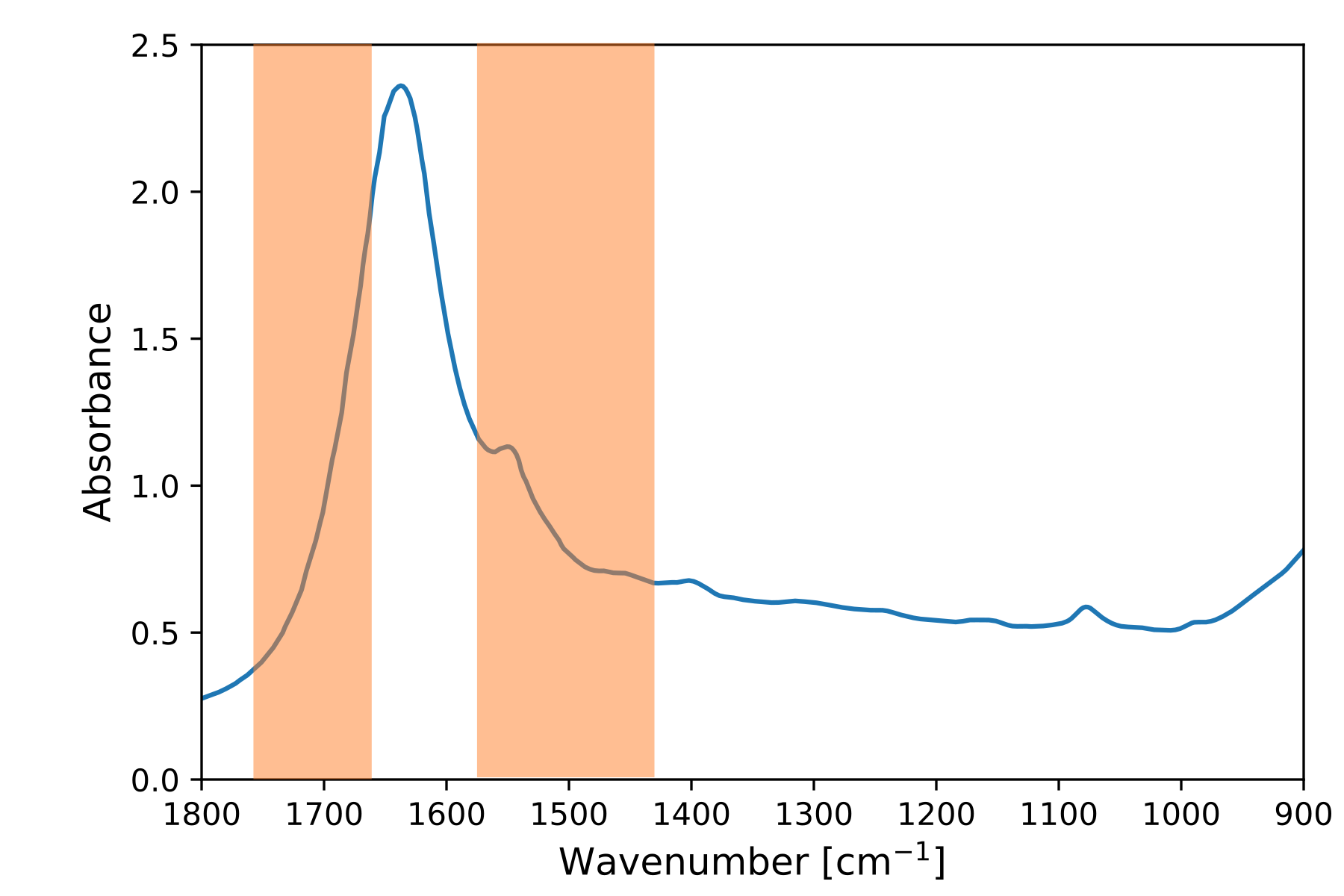}
  \caption{Important regions identified for CNN regression analysis of albumin in the Solutions dataset. }
  \label{fgr:albumin}
\end{figure}

\begin{figure}[h]
\centering
  \includegraphics[height=7cm]{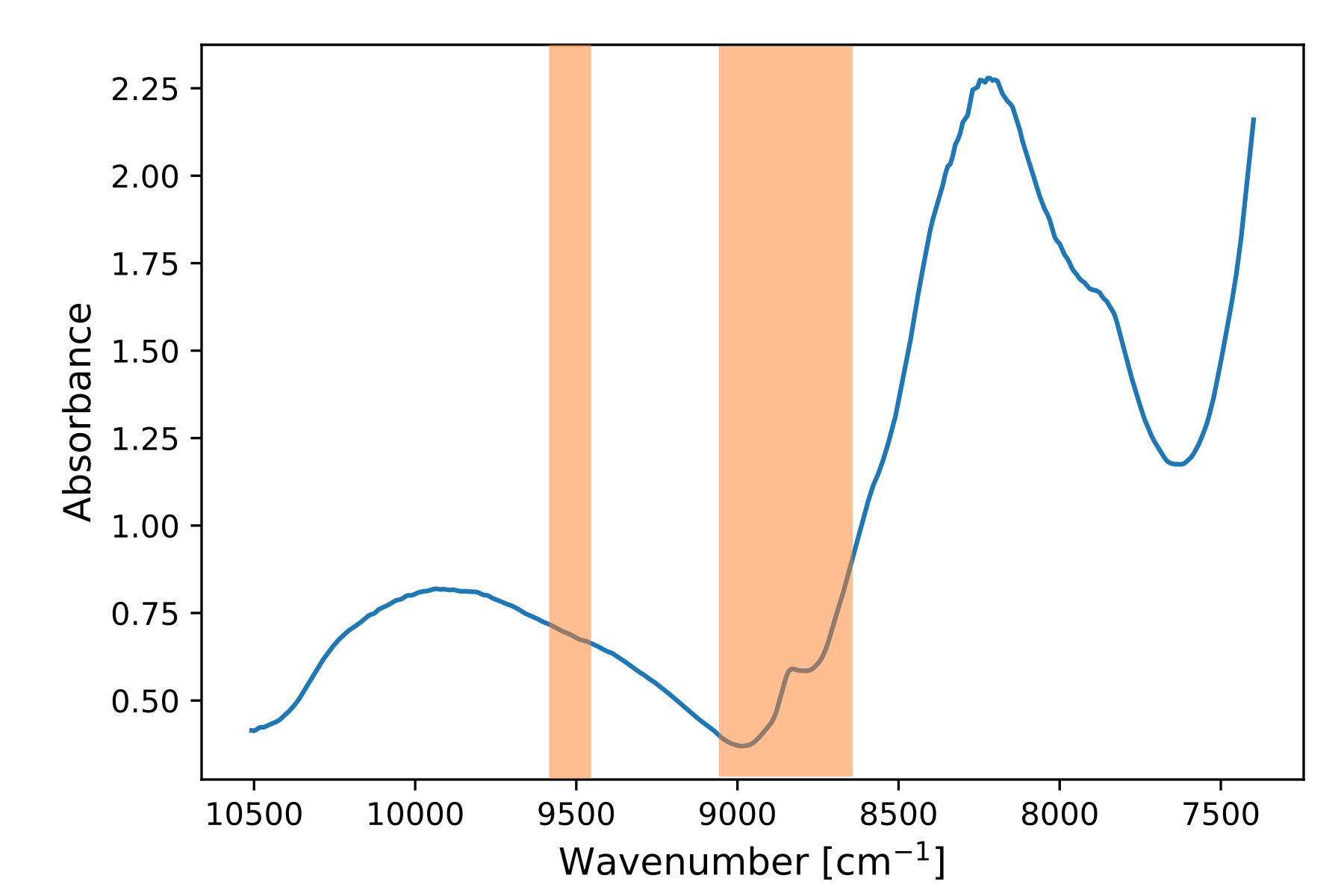}
  \caption{Important regions identified for CNN regression analysis of weight percent in the Tablets NIR dataset. }
  \label{fgr:tablets_nir}
\end{figure}